\documentclass[10pt,twocolumn,letterpaper]{article}

\usepackage[pagenumbers]{cvpr} 

\definecolor{cvprblue}{rgb}{0.21,0.49,0.74}
\usepackage[pagebackref,breaklinks,colorlinks,allcolors=cvprblue]{hyperref}
\usepackage{multirow}
\usepackage{xcolor}
\usepackage{colortbl}

\title{Discriminative Image Generation with Diffusion Models for Zero-Shot Learning}

\author{
    Dingjie Fu$^{1}$, Wenjin Hou$^{2}$, Shiming Chen$^{3}$, Shuhuang Chen$^{1}$, \\
    Xinge You$^{1}\thanks{Corresponding author}$ , Salman Khan$^{3,4}$, Fahad Shahbaz Khan$^{3,5}$ \\
    $^{1}$ Huazhong University of Science and Technology ~~~
    $^{2}$ ReLER Lab, Zhejiang University, China  \\
    $^{3}$ Mohamed bin Zayed University of AI ~~~
    $^{4}$ Australian National University ~~~
    $^{5}$ Linköping University \\
    {\tt\small\{dingjiefu1103,houwj17,gchenshiming\}@gmail.com} \qquad 
}

\begin{document}
\maketitle
\begin{abstract}
Generative Zero-Shot Learning (ZSL) methods synthesize class-related features based on predefined class semantic prototypes, showcasing superior performance. However, this feature generation paradigm falls short of providing interpretable insights. In addition, existing approaches rely on semantic prototypes annotated by human experts, which exhibit a significant limitation in their scalability to generalized scenes. To overcome these deficiencies, a natural solution is to generate images for unseen classes using text prompts. To this end, We present \textbf{DIG-ZSL}, a novel \textbf{D}iscriminative \textbf{I}mage \textbf{G}eneration framework for \textbf{Z}ero-\textbf{S}hot \textbf{L}earning. Specifically, to ensure the generation of discriminative images for training an effective ZSL classifier, we learn a discriminative class token (DCT) for each unseen class under the guidance of a pre-trained category discrimination model (CDM). Harnessing DCTs, we can generate diverse and high-quality images, which serve as informative unseen samples for ZSL tasks. In this paper, the extensive experiments and visualizations on four datasets show that our DIG-ZSL: (1) generates diverse and high-quality images, (2) outperforms previous state-of-the-art nonhuman-annotated semantic prototype-based methods by a large margin, and (3) achieves comparable or better performance than baselines that leverage human-annotated semantic prototypes. The codes will be made available upon acceptance of the paper.
\end{abstract}

\begin{figure}[t]
  \centering
   \includegraphics[width=1.0\linewidth]{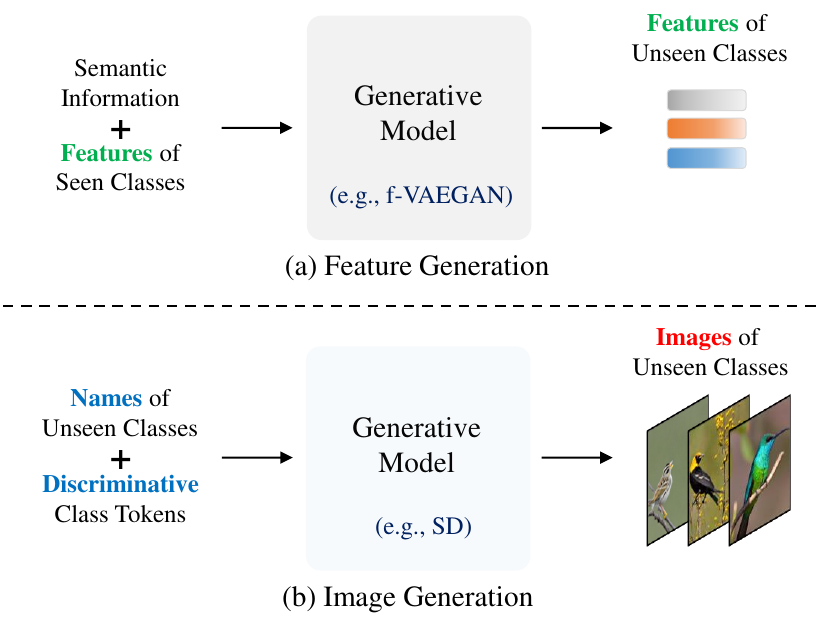}
   \caption{Schematic of the ZSL method. (a) Previous works (\eg, f-VAEGAN \cite{xian2019f}) learn a generative model to \textit{\textbf{synthesize visual features for unseen classes}}, conditioned on human-annotated semantic information and features from seen classes. These approaches results in a lack of interpretable insights. (b) Our proposed DIG-ZSL learns discriminative class tokens for unseen classes to \textit{\textbf{generate images for unseen classes}}, based on a text-to-image diffusion model (\eg, Stable Diffusion (SD) \cite{rombach2022high}) and class names. Our method enables generating diverse and high-quality images that contain discriminative information.}
   \label{fig:motivation}
\end{figure}    
\section{Introduction}
\label{sec:intro}
\quad Zero-Shot Learning (ZSL) aims to train a model capable of classifying images from unseen classes by transferring knowledge from seen classes, utilizing auxiliary semantic information---also known as class semantic prototypes (\eg, attributes \cite{lampert2009learning}, word vectors \cite{socher2013zero}, and textual descriptions \cite{reed2016learning}). Since there are no labeled samples available from unseen classes, generative ZSL methods \cite{xian2019f, narayan2020latent, chen2021hsva, chen2023evolving, hou2024visual} employ generative models such as generative adversarial networks \cite{goodfellow2014generative} and variational autoencoders \cite{Kingma2014AutoEncodingVB} to synthesize data for unseen classes. These methods have showcased the benefits of knowledge transfer and have achieved significant improvements.

However, existing generative ZSL methods \cite{xian2018feature, mishra2018generative, xian2019f, schonfeld2019CADA-VAE, narayan2020latent} mainly focus on the synthesis of visual features, which falls short of providing interpretable insights. As depicted in Fig. \ref{fig:motivation}(a), previous works learn a generative model to synthesize visual features for unseen classes based on their semantic prototypes and features from seen classes. While these methods  achieves high accuracy, they may not be able to offer clear explanations for their synthesized data. In addition, these approaches rely on human-annotated semantic prototypes, exhibiting a significant limitation in their scalability to generalized scenes. 

To overcome these deficiencies, this paper explores the generation of discriminative images for ZSL, leveraging the capabilities of powerful diffusion models (DMs). The advancement of DMs has inspired a variety of innovative applications, including text-to-image generation \cite{rombach2022high, ramesh2022hierarchical}, image inpainting \cite{liu2024structure, corneanu2024latentpaint}, and text-guided image editing \cite{bodur2024iedit, liu2024towards}. Although these methods generate diverse and realistic images, they concentrate solely on evaluation metrics for image generation, neglecting the feasibility of these generated images for classification tasks. A recent study, ZeroDiff \cite{ye2024exploring}, explores the ability of DMs to synthesize informative data. Their experiments demonstrate that incorporating DMs into ZSL is effective and promising. However, this approach still adheres to the established paradigm of feature generation. To enable the generation of informative images, a natural solution lies in the use of advanced text-to-image diffusion models (\eg, Stable Diffusion (SD) \cite{rombach2022high}). Research suggests that combining images produced by SD with real ones enhances the efficacy of supervised learning tasks \cite{he2022synthetic, azizi2023synthetic}. This enhancement is attributed to the enriched dataset diversity and the ability to simulate a broader spectrum of visual scenarios. However, in the ZSL context, where only data from seen classes is available, these generative approaches face significant challenges. They are inherently constrained by the semantic gap between the generated samples and the real-world images. Thus, these methods produce unsatisfactory ZSL performance due to a lack of discriminative information for unseen classes. 

Based upon the aforementioned analysis and findings, we propose a novel \underline{\textbf{D}}iscriminative \underline{\textbf{I}}mage \underline{\textbf{G}}eneration framework for \underline{\textbf{Z}}ero-\underline{\textbf{S}}hot \underline{\textbf{L}}earning, termed \textbf{DIG-ZSL}. As illustrated in Fig. \ref{fig:motivation}(b), DIG-ZSL enables generating diverse and high-quality images for unseen categories using class names and discriminative class tokens. Specifically, we first develop a category discrimination model (CDM) utilizing only the training data from seen classes and nonhuman-annotated semantic prototypes derived from class names (\ie, CLIP text embedding). Then, we optimize a discriminative class token (DCT) for each unseen class, under the guidance of the pre-trained CDM. In the end, we integrate the DCT into a text prompt, thereby transforming it into a discriminative text prompt. Harnessing these prompts, we can generate discriminative images and then train a ZSL classifier to recognize samples from unseen classes. 

In summary, our key contributions are as follows:
\begin{itemize}
\item We take advantage of text-to-image generative models, our DIG-ZSL can generate realistic and discriminative images. To the best of our knowledge, this paper is the first to tackle ZSL in an image generation manner. 
\item We propose learning a discriminative class token for each unseen class under the guidance of a category discrimination model, alleviating the semantic gap between the generated images and the real-world ones.
\item We conduct extensive experiments on AWA2 \cite{xian2018zero}, CUB \cite{welinder2010caltech}, FLO \cite{nilsback2008automated}, and SUN \cite{patterson2012sun} datasets. The comprehensive results show that our DIG-ZSL surpasses the state-of-the-art nonhuman-annotated semantic prototype-based methods with an average Top-1 accuracy improvement of 24.2\%, while remaining competitive with methods that utilize human-annotated semantic prototypes.
\end{itemize}
\begin{figure*}[t]
    \begin{center}
        \includegraphics[width=1.0\linewidth]{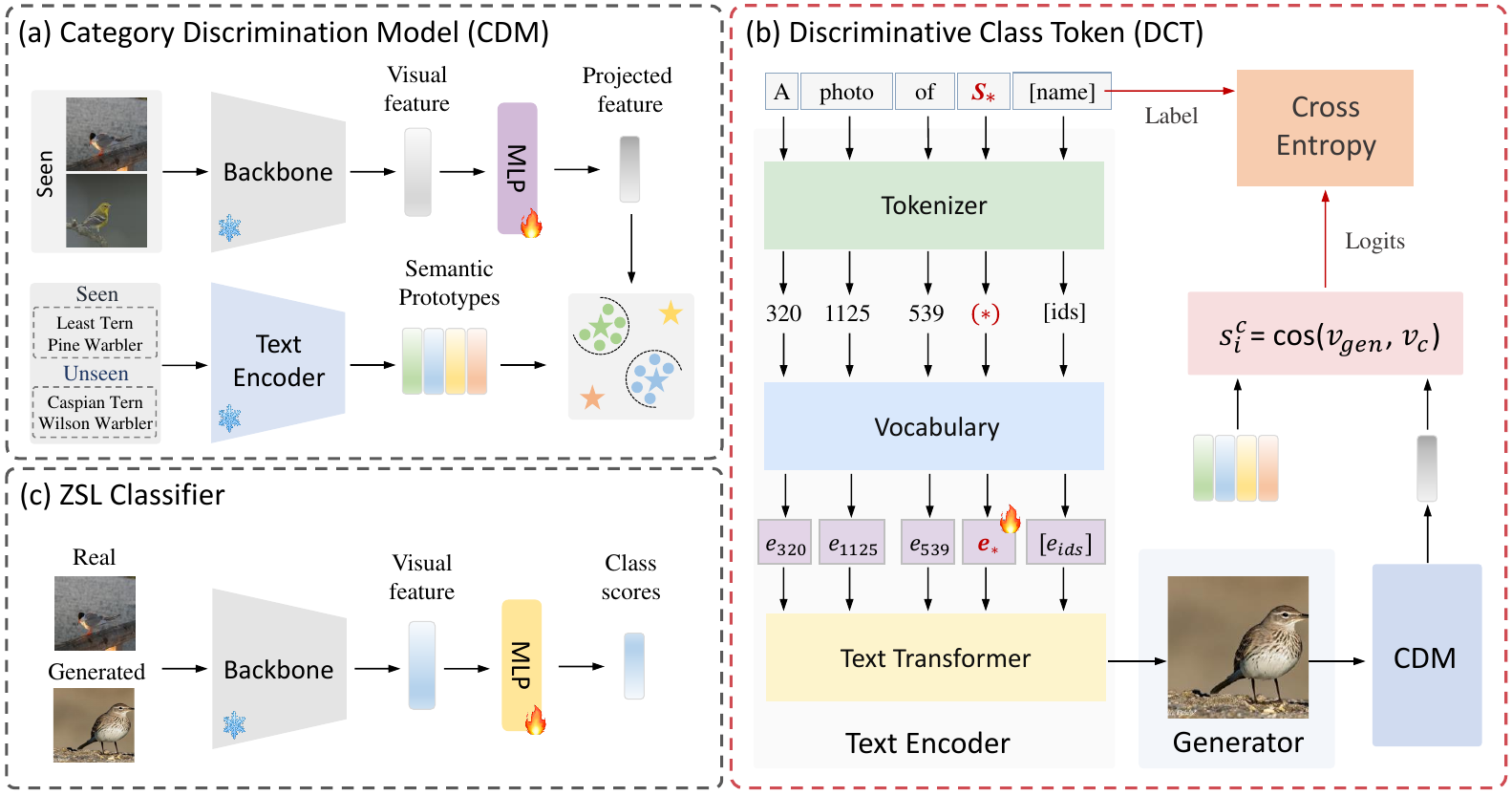}
        \caption{Illustration of our proposed \textbf{DIG-ZSL} framework. We first learn a category discrimination model (CDM) with the training data from seen classes and semantic prototypes derived from class names. Then, we initialize a token $S_*$, and iteratively modifying the embedding of this input token (denoted as $\boldsymbol{e}_*$) under the guidance of CDM, using a cross-entropy loss. Equipped with the optimized token for each unseen category, we incorporate it into the conditioning prompt to generate realistic and discriminative images. Finally, we combine these synthesized data with real training data to learn an effective ZSL classifier.}
        \label{fig:framework}
    \end{center}
    \vspace{-1mm}
\end{figure*}

\section{Related Works}
\label{sec:related}
\noindent\textbf{Zero-Shot Learning.}
ZSL recognizes unseen classes by performing a mapping function between visual and semantic domains, facilitated by semantic prototypes. Existing methods for this task can be categorized into embedding-based and generative-based approaches. Embedding-based methods \cite{xu2020attribute, huynh2020fine, chen2023duet, liu2023progressive, chen2024progressive, chen2024causal} typically project visual features into the semantic space, aligning them with their corresponding semantic prototypes. However, in the case of training only on seen samples, embedding-based methods inevitably run the risk of overfitting to seen classes. To address this issue, generative-based methods \cite{xian2019f, narayan2020latent, chen2021hsva, chen2023evolving, hou2024visual, ye2024exploring} have been introduced to generate samples for unseen classes as a form of data augmentation. The generative ZSL methods have shown notable performance improvements and have recently gained considerable popularity. 

Despite their successes, previous generative-based methods have mainly focused on generating visual features, which limits the interpretability and explainability of the synthesized data. Although f-VAEGAN \cite{xian2019f} shows that the features it learns are interpretable, the process of training an additional visualization model can be cumbersome, and the generated images tend to be low-fidelity. To achieve intuitive and interpretable results, we employ text-to-image diffusion models to generate realistic and discriminative images to advance ZSL.\\
\noindent\textbf{Semantic Prototypes for ZSL.}
Most of the previous generative methods \cite{xian2019f,narayan2020latent,chen2021free, hou2024visual} rely on human-annotated class semantic prototypes, exhibiting a significant limitation in the scalability to generalized scenes. To enable ZSL approaches to be applicable across various domains, several works explore obtaining semantic prototypes without manual annotations \cite{pennington2014glove,beltagy2020longformer, naeem2022i2dformer,naeem2024i2dformer+}. Typically, some methods propose to harness word embeddings from pre-trained language models \cite{socher2013zero,mikolov2013distributed,pennington2014glove} as alternative semantic prototypes, while VGSE \cite{xu2022vgse} learns class embeddings by using image patches and class embedding vectors. Recently, \cite{naeem2022i2dformer,naeem2023i2mvformer,naeem2024i2dformer+} use knowledge from documents and images of seen classes, which demonstrate significant performance improvements. In contrast to these approaches, our DIG-ZSL leverages semantic prototypes extracted from the text encoder of CLIP \cite{radford2021learning}.
We apply these prototypes to train a category discrimination model, which guides the optimization of the discriminative class tokens.\\
\noindent\textbf{Text-to-Image Diffusion Models.}
The field of text-to-image diffusion models has been studied extensively \cite{saharia2022photorealistic, ramesh2022hierarchical, rombach2022high, chang2023muse}. Some methods \cite{tian2024stablerep, tian2024learning} generate multiple images to perform contrastive learning, treating images produced by the same prompt as positive samples. These methods learn representations that exhibit strong transferability to downstream tasks. Additionally, the Diffusion Classifier \cite{li2023your} leverages diffusion models to perform zero-shot classification, achieving satisfactory results on various benchmarks. However, methods such as \cite{tian2024learning} and \cite{tian2024stablerep} require pre-training on extensive images, and \cite{li2023your} consumes a considerable amount of time to classify a single image. Therefore, these methods either demand substantial computational resources or may be unsuitable for downstream tasks that require only a small amount of generated images. 

Our work is closely related to approaches that learn new concepts in the embedding space of a frozen text-to-image model \cite{gal2022image, du2024dream, schwartz2023discriminative}. Textual Inversion \cite{gal2022image} using 3-5 images of a new concept to learn its representation through new \textquotedblleft words\textquotedblright, DREAM-OOD \cite{du2024dream} learns a text-conditioned latent space based on in-distribution data, and \cite{schwartz2023discriminative} optimizes class tokens to resolve ambiguity in the input text and add discriminative features for a given class. However, the method described in \cite{schwartz2023discriminative} assumes that the guidance model has access to data from all classes. In contrast, we utilize a pre-trained category discrimination model that is trained exclusively on the training data from seen classes. 

\section{Methodology} 
\label{sec:methodology}
\quad DIG-ZSL facilitates ZSL by generating realistic and discriminative images for unseen categories. In this section, we first review the definition of the ZSL setting. Then, we proceed to introduce the foundational concepts of Stable Diffusion \cite{rombach2022high}. Subsequently, we elaborate on the detailed design of our proposed DIG-ZSL framework, as illustrated in Fig. \ref{fig:framework}. At the end of this section, we demonstrate the discriminative class token training strategy and outline the process for zero-shot prediction.
\subsection{Preliminaries}
\label{sec:preliminary}
\noindent\textbf{Problem Definition.}
Assume that the data of seen classes $\mathcal{D}^{s}=\left\{\left(x_{i}^{s}, y_{i}^{s}\right)\right\}$ comprises $\mathcal{C}^s$ classes, where $x_i^s \in \mathcal{X}$ and $y_i^s \in \mathcal{Y}^s$ represent the \textit{i}-th sample and its corresponding class label, respectively. $\mathcal{D}^{s}$ is further divided into a training set $\mathcal{D}^{s}_{tr}$ and a test set $\mathcal{D}^{s}_{te}$. A disjoint set $\mathcal{D}^{u}=\left\{\left(x_{i}^{u}, y_{i}^{u}\right)\right\}$ consists of $\mathcal{C}^u$ unseen classes, where $x_{i}^{u}\in \mathcal{X}$ are the unseen class samples, and $y_{i}^{u} \in \mathcal{Y}^u$ are the corresponding labels. In addition, auxiliary semantic information is available for all classes $c \in \mathcal{C}^{s} \cup \mathcal{C}^{u} = \mathcal{C}$, which facilitates the transfer of knowledge from seen classes to unseen classes. In this paper, we leverage the semantic prototype of the class $c$ extracted from the text encoder of CLIP \cite{radford2021learning}, denoted as $\boldsymbol{v}_{c}$. In the conventional zero-shot learning (CZSL) setting, the primary objective is to classify test images that belong to unseen classes. In the generalized zero-shot learning (GZSL) setting, our aim is to develop a classifier capable of recognizing both seen and unseen classes.\\
\noindent\textbf{Stable Difusion.}
Stable Difusion (SD) \cite{rombach2022high} is a denoising diffusion probabilistic model \cite{sohl2015deep, ho2020denoising} designed to learn a data distribution $\mathcal{E}(x)$ in the latent space. During training, it initiates with input latent data $z_0\sim \mathcal{E}(x)$ and progressively generates a noisy version $z_t$ over a total of $T$ steps, which is formulated as:
\begin{equation}
\label{eq:ddpm_forward2}
z_t = \sqrt{\overline{\alpha}_t} \cdot z_0 + \sqrt{1 - \overline{\alpha}_t} \cdot \epsilon,
\end{equation}
where $\overline{\alpha}_t$ denotes the predetermined schedule with respect to the diffusion step $t$, and $\epsilon \sim \mathcal{N}(0,\text{I})$ represents the added noise. With the forward process defined in Eq. \ref{eq:ddpm_forward2}, the model is trained to predict added noise using a denoising autoencoder $\epsilon_\theta\big(z_t, t, \mathcal{T}(p)\big)$. The loss of the conditional denoising process is given by:
\begin{equation}
\label{eq:loss_cond}
\mathcal{L}_{cond} = \mathbb{E}_{z,p,\epsilon,t}\big[||\epsilon -  \epsilon_\theta\big(z_t, t, \mathcal{T}(p)\big)||^2_2\big],
\end{equation}
where $\mathcal{T}(\cdot)$ denotes the text encoder of the SD, and $p$ refers to text prompt \cite{rombach2022high} in the text-to-image pipeline.

The images generated by SD suffer the effects of the hyper-parameters. It improves the sample quality and the text-image alignment with a higher classifier-free guidance (CFG) scale \cite{ho2022classifier}, which combines conditional score estimate $\hat{\epsilon}_\theta\big(z_t, \mathcal{T}(p)\big)$ and unconditional estimate $\hat{\epsilon}_\theta\big(z_t)$ with the guidance scale $w$, denoted as:
\begin{equation}
\label{eq:cfg}
\tilde{\epsilon}_\theta\big(z_t, \mathcal{T}(p)\big) = w\hat{\epsilon}_\theta\big(z_t, \mathcal{T}(p)\big) + (1-w)\hat{\epsilon}_\theta(z_t),
\end{equation}
where $\hat{\epsilon}_\theta\big(z_t)$ is obtained by using an empty input prompt.
\subsection{Category Discrimination Model}
\label{sec:cdm}
\quad To guide the optimization process of the discriminative class token for each unseen class, we initially train a category discrimination model (CDM). Fig. \ref{fig:framework}(a) depicts the architecture of the CDM. We employ a frozen backbone (\ie, ViT \cite{dosovitskiy2020ViT}), to extract the visual feature $\boldsymbol{f}_i\in \mathbb{R}^{d_v\times{1}}$ of the image $x_i$. Then, we deploy a trainable multi-layer perceptron (MLP) to map the visual feature to the semantic space, obtaining the projected feature $\boldsymbol{a}_i\in \mathbb{R}^{d_t\times{1}}$ for visual-semantic alignment. Here, $d_v$ and $d_t$ represent the dimension of the visual feature and the projected feature, respectively. Using class semantic prototypes derived from the text encoder of CLIP \cite{radford2021learning}, the CDM facilitates the recognition of images from unseen classes.

Specifically, the CDM is trained exclusively on the training data from seen classes and is designed to guide generated images towards the unseen classes corresponding to their text prompts. Thanks to the CDM's ability to perform category discrimination, the distribution of the generated images shows greater similarity to that of real images from unseen classes. Moreover, we eliminate the use of human-annotated attributes, which are widely used in mainstream methods. While we train our own CDM, our framework is also compatible with off-the-shelf models.
\subsection{Discriminative Class Token}
\label{sec:dctl}
\quad To generate realistic and informative images that facilitate the training of an effective ZSL classifier, we propose learning a discriminative class token (DCT) for each unseen class. Drawing inspiration from previous work \cite{schwartz2023discriminative}, our main idea involves using a pre-trained CDM to guide this process. Contrary to the guidance model in \cite{schwartz2023discriminative}, which is trained on data encompassing all categories, our CDM accesses only the training data from seen categories. Therefore, our approach adheres to the ZSL setting and ensures a fair comparison with other ZSL competitors.

In Fig. \ref{fig:framework}(b), we illustrate the optimization of DCTs, which involves iteratively modifying the embedding of an added input token (\ie, $\boldsymbol{e}_*$) in a text-to-image diffusion model. Specifically, we first introduce a discriminative token, denoted as $S_*$, which is not present in the vocabulary of the text encoder. We then construct a custom text prompt by integrating a prompt prefix (\eg, \textquotedblleft A photo of\textquotedblright), the discriminative token, and the target class name, obtaining a discriminative text prompt $p=$ \textquotedblleft A photo of $S_*$ [name]\textquotedblright. Given the input prompt, a tokenizer converts it into tokens, which are indices of single words in the embedding vocabulary. For each token, its corresponding embedding vector $\boldsymbol{e}$ is retrieved through the index-based lookup scheme. The text transformer processes these embedding vectors into an embedding representation $\mathcal{T}(p)$, which facilitates the conditional image generation.

After generating an image $\widetilde{x}_i$ from the generator $\mathcal{G}$, we utilize the pre-trained CDM to extract its semantic feature, denoted as $\boldsymbol{v}_{gen}$. Since the CDM only accesses the data from seen classes, it cannot assign a probability score to each unseen category. Adhering to embedding-based approaches in ZSL, we propose to compute the class score by calculating cosine similarity between $\boldsymbol{v}_{gen}$ and $\boldsymbol{v}_{c}$ for classification, formulated as:
\begin{equation}
\label{eq:simi}
s^c_i = \text{cos}(\boldsymbol{v}_{gen}, \boldsymbol{v}_{c}),
\end{equation}
where cos(, ) refers to the cosine similarity function.
Accordingly, we optimize $\boldsymbol{e}_*$ using a cross-entropy loss, defined as follows:
\begin{equation}
\label{eq:celoss}
\mathcal{L}_{ce} = -\mathbb{E}\big[\log\frac{\text{exp} \ (s^c_i)}{\sum_{c'\in\mathcal{C}^u}{\text{exp} \ (s^{c'}_i)}}\big].
\end{equation}

In this paper, we learn DCTs for unseen categories. Thus, Eq. (\ref{eq:celoss}) is designed to distinguish only among different unseen classes.
\subsection{ZSL Classifier}
\quad Once we have obtained the DCT for each unseen class, we incorporate it into a text prompt to generate realistic and informative images, denoted as $\widetilde{\mathcal{D}}^{u}=\left\{\left(\widetilde{x}_{i}^{u}, \widetilde{y}_{i}^{u}\right)\right\}$. As shown in Fig. \ref{fig:framework}(c), we employ the same backbone as the CDM to extract latent features. Consequently, we train an additional MLP, \ie, $f_{czsl} : \mathcal{X} \rightarrow \mathcal{Y}^u$ in the CZSL setting, and $f_{gzsl} : \mathcal{X} \rightarrow \mathcal{Y}^s\cup\mathcal{Y}^u$ in the GZSL setting, respectively.
\subsection{Training and Inference}
\label{sec:ta}
\noindent\textbf{Overview.}
Our DIG-ZSL framework adopts a three-stage approach. In the initial stage, we train a category discrimination model to provide discriminative signals for unseen categories. Then, we optimize discriminative class tokens for unseen classes to produce realistic and informative images. Finally, we perform zero-shot classification by combining the generated images with real ones. In the following, we introduce the optimization strategy for DCT learning and illustrate how we perform zero-shot prediction. \\
\noindent\textbf{Optimization Strategy.}
Our DIG-ZSL eliminates the requirement for training on synthesized data, leading the generated images towards a given target class based on the discriminative signals from the CDM. However, guidance signals that deviate from the real data distribution may lead to a decrease in performance. Therefore, we aim to regulate the \textquotedblleft training intensity\textquotedblright \ considering three properties: (1) initializing the DCT as the placeholder token \textquotedblleft a\textquotedblright, (2) setting a threshold $\gamma$ informed by the performance of our CDM, and (3) employing an early stopping strategy to mitigate over-training. More specifically, while the generated images achieve a classification accuracy of $\gamma$ using the CDM, we cease further optimization of this DCT.

\noindent\textbf{ZSL Inference.}
we perform inference using the test sets $\mathcal{D}^{s}_{te}$ and $\mathcal{D}^{u}$. Since generating numerous images per class is time-consuming and resource-intensive, we set a fixed number of generated images per unseen class across all datasets. Therefore, in order to alleviate the tendency of overfitting to seen classes, we set a coefficient $\lambda$ to calibrate the predictor sensitivity to unseen classes. To this end, we predict the class label $c^*$ using the following formulation:
\begin{equation}
\label{eq:inference}
c^* = \text{argmax}_{c\in \mathcal{C}} (o - \lambda\mathbb{I}_{[c\in\mathcal{C}^s]}),
\end{equation}
where $o$ denotes the output logits after softmax operation, and $\mathbb{I}_{[c\in\mathcal{C}^s]}$ is an indicator function, which is 1 when $c\in\mathcal{C}^s$ and 0 otherwise.
\section{Experiments}
\label{sec:experiments}
\subsection{Experimental Configurations}
\noindent\textbf{Benchmark Datasets.}
We conduct extensive experiments on four widely used ZSL benchmark datasets, \ie, Animals with Attributes 2 (AWA2) \cite{xian2018zero}, Caltech-USCD Birds-200-2011 (CUB) \cite{welinder2010caltech}, Oxford Flowers (FLO) \cite{nilsback2008automated}, and SUN Attribute (SUN) \cite{patterson2012sun}. Following the unified evaluation protocols provided by \cite{xian2018zero}. All datasets are split into seen and unseen classes. In AWA2, there are 37,322 images and 50 animal classes, where 40 are seen and 10 are unseen. In CUB, the image number, bird class number, and seen/unseen class numbers are 11,788, 200, and 150/50, respectively. In SUN, these numbers are 14,340, 717, and 645/72, respectively. In FLO, these numbers are 8,189, 102, and 82/20, respectively. These configurations are the same for all the methods. 

\noindent\textbf{Evaluation Metrics.}
Following previous work \cite{xian2018zero}, the performance is evaluated by the Top-1 accuracy of the unseen class (denoted as $acc$) for the CZSL setting. In the GZSL setting, we measure the harmonic mean between seen and unseen classes: $H = (2\times{S}\times{U}) / (S+U)$, where $S$ denotes the accuracy of seen classes and $U$ denotes the accuracy of unseen classes.

\begin{table}[htbp]
  \centering
    \caption{Comparison between our DIG-ZSL and existing methods that leverage nonhuman-annotated semantic prototypes on four datasets under CZSL setting. Our DIG-ZSL significantly improves on the baselines. Best results are in \textbf{boldface}.}
  \vspace{-1mm}
      \begin{tabular}{l|c|c|c|c}
        \hline
        \multirow{2}{*}{\textbf{Methods}} & \textbf{AWA2} & \textbf{CUB} & \textbf{FLO} & \textbf{SUN} \\
        \cline{2-5}
        & acc & acc & acc & acc \\
        \hline
        Glove \cite{pennington2014glove}  & 52.1 & 20.4 & 21.6 & -- \\
        LongFormer \cite{beltagy2020longformer}  & 44.2 & 22.6 & 8.8 & -- \\
        MPNet \cite{song2020mpnet}  & 61.8 & 25.8 & 26.3 & -- \\
        TF-IDF \cite{salton1988term}  & 46.4 & 39.9 & 34.0 & -- \\
        VGSE-APN \cite{xu2022vgse}  & 64.0 & 28.9 & -- & 38.1 \\
        I2DFormer \cite{naeem2022i2dformer} & 76.4 & 45.4 & 40.0 & -- \\
        I2MVFormer \cite{naeem2023i2mvformer} & 73.6 & 42.1 & 41.3 & -- \\
        I2DFormer+ \cite{naeem2024i2dformer+} & 77.3 & 45.9 & 41.3 & -- \\
        \hline
        \rowcolor{blue!5}
        DIG-ZSL (Ours)  & \textbf{90.1} & \textbf{69.7} & \textbf{70.9} & \textbf{68.8} \\
        \hline
    \end{tabular}
      \label{tab:weak_czsl}
       \vspace{-1mm}
\end{table}

\begin{table*}[t]
\centering  
\caption{Comparing DIG-ZSL with nonhuman-annotated semantic prototype-based methods in the GZSL setting. We report the Top-1 accuracy on seen/unseen (S/U) classes and their harmonic mean (H). We see that DIG-ZSL outperforms the previous SOTA results of I2MVFormer-LLM. The results of large pre-trained vision-language models are in \textcolor{gray}{gray}. Best results are highlighted in \textbf{boldface}.}
\resizebox{\linewidth}{!}{
    \begin{tabular}{l|c|ccc|ccc|ccc|ccc}
    \hline
        \multirow{2}{*}{\textbf{Methods}} &\multirow{2}{*}{\textbf{Venue}} 
        &\multicolumn{3}{c|}{\textbf{AWA2}}
        &\multicolumn{3}{c|}{\textbf{CUB}} 
        &\multicolumn{3}{c|}{\textbf{FLO}} 
        &\multicolumn{3}{c}{\textbf{SUN}}\\
        \cline{3-14}
        &&\rm{U}&\rm{S}&\rm{H} 
        &\rm{U}&\rm{S}&\rm{H} 
        &\rm{U}&\rm{S}&\rm{H}
        &\rm{U}&\rm{S}&\rm{H}\\
        \hline
        MPNet \cite{song2020mpnet}& NeurIPS'20
        & 58.0 & 76.4 & 66.0
        & 20.6 & 44.3 & 28.2
        & 22.2 & 96.7 & 36.1
        & -- & -- & -- \\
        VGSE-APN \cite{xu2022vgse} & CVPR'22 
        & 51.2 & 81.8 & 63.0 
        & 21.9 & 45.5 & 29.5 
        & -- & -- & -- 
        &24.1 & 31.8 & 27.4\\
        I2DFormer \cite{naeem2022i2dformer} & NeurIPS'22
        & 66.8 & 76.8 & 71.5 
        & 35.3 & 57.6 & 43.8 
        & 35.8 & 91.9 & 51.5 
        & -- & -- & -- \\
        I2MVFormer-Wiki \cite{naeem2023i2mvformer} & CVPR'23 
        & 66.6 & 82.9 & 73.8 
        & 32.4 & 63.1 & 42.8 
        & 34.9 & 96.1 & 51.2 
        & -- & -- & -- \\
        I2MVFormer-LLM \cite{naeem2023i2mvformer} & CVPR'23 
        & 72.7 & 81.3 & 76.8 
        & 40.1 & 58.0 & 47.4 
        & 41.1 & 91.1 & 56.6 
        & -- & -- & -- \\
        TF-VAEGAN+SHIP \cite{wang2023SHIP} & ICCV'23 
        & 43.7 & \textbf{96.3} & 60.1 
        & 21.1 & \textbf{84.4} & 34.0 
        & 37.4 & \textbf{97.2} & 54.0 
        & -- & -- & -- \\
        I2DFormer+ \cite{naeem2024i2dformer+} & IJCV'24 
        & 69.8 & 83.2 & 75.9 
        & 38.3 & 55.2 & 45.3
        & 36.9 & 86.9 & 51.8 
        & -- & -- & -- \\
        \hline
        \textcolor{gray}{CLIP \cite{radford2021learning}}& \textcolor{gray}{ICML'21}
        & \textcolor{gray}{--} & \textcolor{gray}{--} & \textcolor{gray}{--}
        & \textcolor{gray}{55.2} & \textcolor{gray}{54.8} & \textcolor{gray}{55.0}
        & \textcolor{gray}{65.6} & \textcolor{gray}{67.9} & \textcolor{gray}{66.7}
        & \textcolor{gray}{--} & \textcolor{gray}{--} & \textcolor{gray}{--} \\
        \textcolor{gray}{CoOp \cite{zhou2022CoOp}}& \textcolor{gray}{IJCV'22}
        & \textcolor{gray}{72.7} & \textcolor{gray}{95.3} & \textcolor{gray}{82.5}
        & \textcolor{gray}{49.2} & \textcolor{gray}{63.8} & \textcolor{gray}{55.6}
        & \textcolor{gray}{52.2} & \textcolor{gray}{85.8} & \textcolor{gray}{64.9}
        & \textcolor{gray}{--} & \textcolor{gray}{--} & \textcolor{gray}{--} \\
        \hline
        \rowcolor{blue!5}
        DIG-ZSL (Ours) & -- 
        & \textbf{83.9} & 85.8 & \textbf{84.9}
        & \textbf{59.1} & 68.3 & \textbf{63.3}
        & \textbf{70.8} & 96.7 & \textbf{81.7}
        & \textbf{53.5} & \textbf{44.4} & \textbf{48.5} \\
        \hline
    \end{tabular}}
\label{tab:weak_gzsl}
\end{table*}

\begin{table}[htbp]
  \centering
    \caption{Comparing our DIG-ZSL with human-annotated semantic prototype-based methods. We compare with both embedding and generative ZSL methods in the CZSL setting. The best results are marked in \textbf{boldface}.}
  \vspace{-1mm}
  \scalebox{1.0}{
      \begin{tabular}{c|l|c|c|c}
        \hline
        & \multirow{2}{*}{\textbf{Methods}}& \textbf{AWA2} & \textbf{SUN} & \textbf{CUB} \\
        \cline{3-5} && acc & acc & acc \\
        \hline
        \multirow{5}{*}{{\rotatebox{90}{Embedding}}}
        & PREN \cite{ye2019PRNE} & 74.1 & 62.9 & 66.4 \\
        & APN \cite{xu2020APN} & 68.4 & 61.6 & 72.0 \\
        & TCN \cite{jiang2019TCN} & 71.2 & 61.5 & 59.5 \\
        & DAZLE \cite{huynh2020fine}  & 67.9 & 59.4 & 66.0 \\
        & DUET \cite{chen2023duet} & 69.9 & 64.4 & \textbf{72.3} \\
        \hline
        \multirow{7}{*}{{\rotatebox{90}{Generative}}}
        &f-CLSWGAN \cite{xian2018feature} & 68.2 & 60.8 & 57.3 \\
        &f-VAEGAN \cite{xian2019f} & 71.1 & 64.7 & 61.0 \\
        &CADA-VAE \cite{schonfeld2019CADA-VAE} & 63.0 & 61.7 & 59.8 \\
        &TF-VAEGAN \cite{narayan2020latent} & 72.2 & 66.0 & 64.9\\
        &HSVA \cite{chen2021hsva} & 70.6 & 63.8 & 62.8\\
        &f-VAEGAN+DSP \cite{chen2023evolving} & 71.6 & 68.6 & 62.8\\
        \cline{2-5}
        & \cellcolor{blue!5}DIG-ZSL (Ours)
        & \cellcolor{blue!5}\textbf{90.1} & \cellcolor{blue!5}\textbf{68.8} & \cellcolor{blue!5}69.7 \\
        \hline
    \end{tabular}}
      \label{tab:SOTA}
       \vspace{-1mm}
\end{table}

\noindent{\textbf{Implementation Details.}}
We extend the first image generation method to advance zero-shot learning. For image generation, we apply Stable-Diffusion-v2.1 \cite{rombach2022high} to produce 100 samples for each unseen class across all datasets. In the category discrimination model, class semantic protoypes with 512 dimensions are extracted from the text encoder of CLIP ViT-B/16 \cite{radford2021learning}. We leverage the input prompt \textquotedblleft A photo of a [name]\textquotedblright ~to obtain these semantic prototypes. We empirically set the threshold and calibration coefficient \{$\gamma$, $\lambda$\} to \{0.4, 0.95\}, \{0.6, 0.6\}, \{0.4, 0.9\} and \{0.6, 0.8\} for CUB, SUN, FLO, and AWA2, respectively. We train all models on a single NVIDIA 3090 GPU and implement our experiments in PyTorch. Due to page limitations, more configurations are detailed in Appendix \ref{sec:appendA}.

\subsection{Comparison with State-of-the-Arts}
\noindent\textbf{Nonhuman-Annotated Semantic Prototypes.}
We first compare our proposed DIG-ZSL with state-of-the-art (SOTA) methods that make use of nonhuman-annotated semantic prototypes, such as word vectors \cite{pennington2014glove} and text descriptions \cite{naeem2022i2dformer, naeem2024i2dformer+}. Tab. \ref{tab:weak_czsl} shows the results on four benchmark datasets. Our DIG-ZSL achieves the highest Top-1 accuracy values of 90.1\%, 69.7\%, 70.9\% and 68.8\% on AWA2, CUB, FLO, and SUN, respectively. Specifically, on these datasets, DIG-ZSL exceeds the previous SOTA I2DFormer+ \cite{naeem2024i2dformer+} and VGSE \cite{xu2022vgse} by at least 12.8\%, 23.8\%, 29.6\%, and 30.7\%. Overall, these results demonstrate the superior effectiveness and efficiency of our DIG-ZSL when compared to approaches reliant on nonhuman-annotated semantic prototypes in the CZSL setting. 

We also compare our DIG-ZSL with the SOTA methods in the GZSL setting, as the results shown in Tab. \ref{tab:weak_gzsl}. DIG-ZSL gives the best results of $H$$=$84.9\%, $H$$=$63.3\%, $H$$=$81.7\%, $H$$=$48.5\% on AWA2, CUB, FLO, and SUN, respectively. Specifically, on FLO and SUN datasets, our approach increases performance from 56.6\% and 27.4\% to 81.7\% and 48.5\%. On the coarse-grained dataset AWA2, DIG-ZSL outperforms I2MVFormer-LLM with an accuracy improvement of 8.1\%. On the challenging dataset CUB, DIG-ZSL obtains a performance improvement of 18.0\%. In particular, TFVAEGAN + SHIP \cite{wang2023SHIP} using the CLIP Encoder \cite{radford2021learning} achieves the best accuracy for seen classes, but fails to generalize well to unseen classes. The I2MVFormer-LLM \cite{naeem2023i2mvformer} employs large language models (LLMs) to synthesize multi-view knowledge of a class, yet it still does not match the performance of our DIG-ZSL. Notably, as indicated in gray font, our DIG-ZSL also significantly outperforms the large-scale vision-language based ZSL methods (\ie, CLIP \cite{radford2021learning} and CoOp \cite{zhou2022CoOp}). These results indicate that our DIG-ZSL is beneficial for coarse- and fine-grained tasks and demonstrates competitiveness with methods that leverage CLIP or LLMs.

\vspace{-0.01mm}
\noindent\textbf{Human-Annotated Semantic Prototypes.}
We further extend the comparison of our DIG-ZSL with methods that leverage human-annotated semantic prototypes, \ie human-annotated attributes. The results are presented in Tab. \ref{tab:SOTA}. The proposed DIG-ZSL achieves the best Top-1 accuracy on AWA2 and SUN datasets, \eg, 90.1\% and 68.8\%, respectively. Especially, our DIG-ZSL outperforms the recent methods by at least 16.0\% on AWA2. For the scene dataset SUN, DIG-ZSL maintains competitive with an accuracy of 68.8\%. However, for the CUB dataset, DIG-ZSL encounters challenges ($acc$$=$69.7\%). This is attributed to difficulties in generating complex and fine-grained images. These experiments illustrate that DIG-ZSL achieves satisfactory results when dealing with coarse-grained or simpler tasks, but further improvement is needed for challenging fine-grained datasets. Notably, our DIG-ZSL relies solely on semantic prototypes derived from class names, which indicates that DIG-ZSL is more adaptable to generalized scenes. These results consistently demonstrate that our DIG-ZSL is a desirable generation paradigm for ZSL.

\subsection{In-Depth Analysis}

\noindent\textbf{FID Evaluation.}
We employ the Fréchet Inception Distance (FID) \cite{heusel2017gans} score to assess the quality of the generated images of unseen categories in comparison to the real datasets, as shown in the Tab. \ref{tab:fid}. For both methods, we generate 100 images per unseen class and compute the FID score against real unseen images. Our DIG-ZSL outperforms the vanilla SD when leveraging discriminative class tokens. This comparison demonstrates that DIG-ZSL enhances the quality of the generated samples.

\begin{table}[htbp]
  \centering
    \caption{FID score for generated images with SD and our DIG-ZSL. We present results on CUB and SUN datasets. Our approach generates higher-quality images compared to SD.}
  \vspace{-1mm}
      \begin{tabular}{c|c|c}
        \hline
        Datasets & Method & FID$\downarrow$ \\
        \hline
        \multirow{2}{*}{\textbf{CUB}}  & SD & 19.7  \\
        & \cellcolor{blue!5}DIG-ZSL & \cellcolor{blue!5}\textbf{19.6} \\
        \hline
        \multirow{2}{*}{\textbf{SUN}}  & SD & 50.5   \\
        & \cellcolor{blue!5}DIG-ZSL & \cellcolor{blue!5}\textbf{49.9} \\
        \hline
    \end{tabular}
      \label{tab:fid}
       \vspace{1mm}
\end{table}

\noindent\textbf{The impact of CDM.}
We further investigate the impact of CDM on ZSL performance. Tab. \ref{tab:cdm} presents the performance of our DIG-ZSL approach when using different CDMs. We observe that the performance of DIG-ZSL is significantly enhanced when the CDM is fine-tuned on data from seen classes. It should be noted that we only incorporate the backbone into the CDM training process, while keeping the backbone frozen during the training of the ZSL classifier. Specifically, DIG-ZSL$^{\dagger}$ leads to superior accuracy. In terms of $acc$, we observe varying degrees of performance improvement,  with a value of 1.2\%, 2.0\%, 5.1\% and 4.0\% on AWA2, CUB, FLO and SUN, respectively. In general, the consistent improvement validates the effectiveness of a powerful CDM.

\begin{table}[htbp]
  \centering
    \caption{Studies on category discrimination models with different performance. We report the Top-1 accuracy in CZSL setting. The symbol ${\dagger}$ denotes using a fine-tuned CDM.}
  \vspace{-1mm}
      \begin{tabular}{l|c|c|c|c}
        \hline
        \multirow{2}{*}{\textbf{Methods}} & \textbf{AWA2} & \textbf{CUB} & \textbf{FLO} & \textbf{SUN} \\
        \cline{2-5}
        & acc & acc & acc & acc \\
        \hline
        DIG-ZSL & 90.1 & 69.7 & 70.9 & 68.8 \\
        DIG-ZSL$^{\dagger}$ & 91.3$^{\color{purple}\text{+1.2}}$ & 71.7$^{\color{purple}\text{+2.0}}$ & 76.0$^{\color{purple}\text{+5.1}}$ & 72.8$^{\color{purple}\text{+4.0}}$ \\
        \hline

    \end{tabular}
      \label{tab:cdm}
       \vspace{1mm}
\end{table}

\subsection{Qualitative Results}
\begin{figure}[t]
    \begin{center}
        \includegraphics[width=0.9\linewidth]{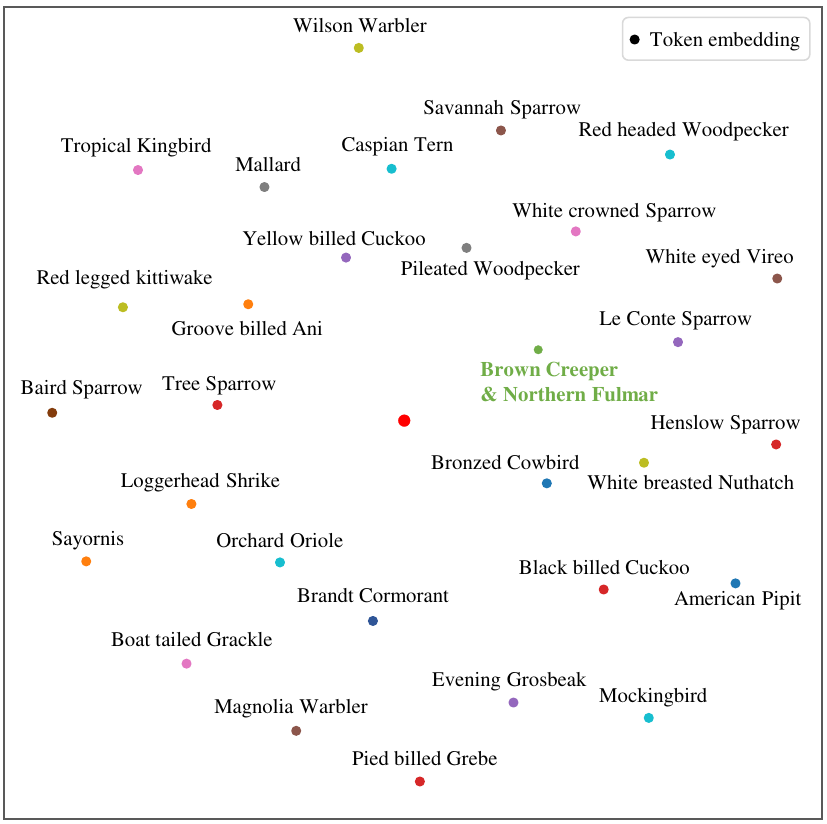}
        \caption{t-SNE visualization of the DCT embeddings. A dot indicates a DCT embedding for an unseen category from CUB dataset. \textcolor{Red}{\textbf{Red}} dot and \textcolor{Green}{\textbf{green}} dot indicate the tokens that are clustered together. (Please zoom in for details.)}
        \label{fig:dct_e}
    \end{center}
    \vspace{-1mm}
\end{figure}

\noindent\textbf{Visualization of the DCT Embeddings.}
We further present the t-SNE visualization of the learned DCT embeddings for CUB dataset. As depicted in Fig. \ref{fig:dct_e}, the embeddings appear to be scattered and discriminative from one another. Given that we initialized the DCT as \textquotedblleft a\textquotedblright, there are cases when the DCT reaches the threshold $\gamma$ without undergoing training, resulting in no optimization for this token. These tokens are more cohesive, as indicated by the red dot in the figure. Similarly, two class tokens (\ie, Brown Creeper, and Northern Fulmar) are observed to be clustered together, as highlighted by the green dot. These findings confirm the promise of our DIG-ZSL in advancing ZSL by providing discriminative token embeddings.

\begin{figure*}[t]
  \centering
   \includegraphics[width=1.0\linewidth]{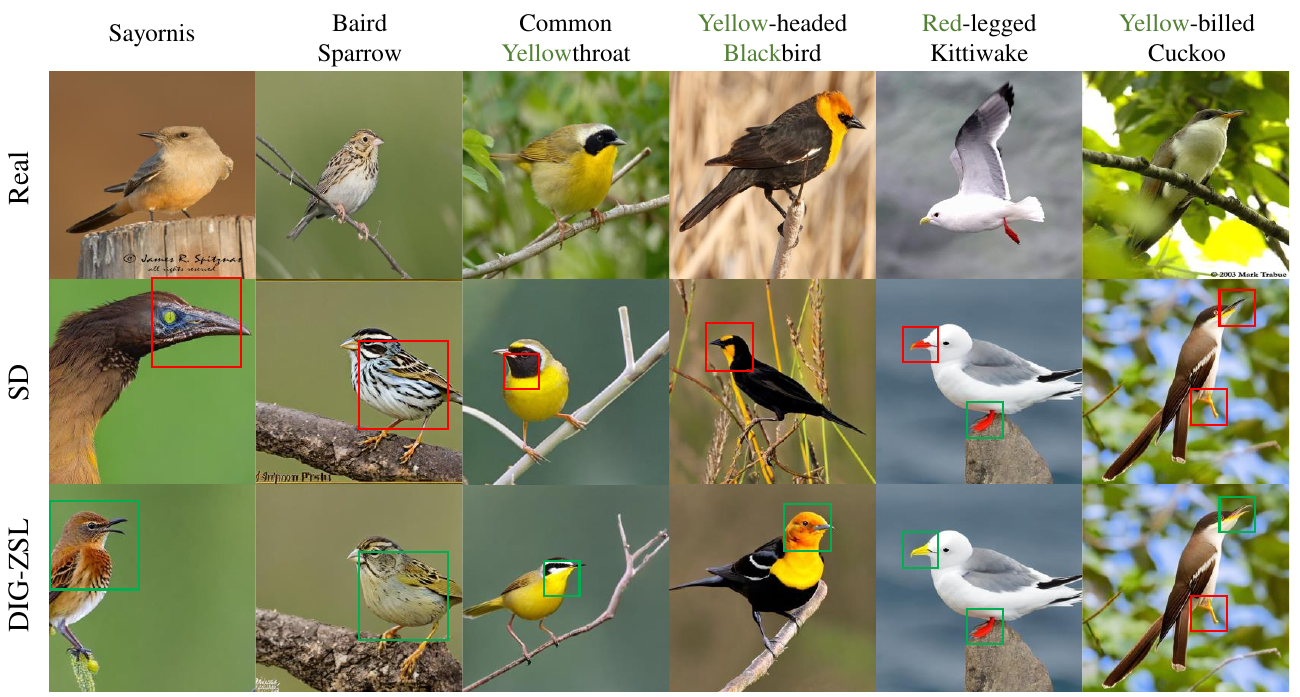}
   \caption{Visualization of generated images \textbf{at the identical step} for both the plain Stable Diffusion (SD) and DIG-ZSL. Our DIG-ZSL demonstrates the capability to enhance text-to-image alignment and significantly improve the portrayal of details. We use CUB as an example, with the \textcolor{Green}{\textbf{green}} box indicating correct details, while the \textcolor{Red}{\textbf{red}} box represents incorrect details.}
   \label{fig:compare_imgs}
\end{figure*}

\begin{figure}[htbp]
    \begin{center}
        \includegraphics[width=1.0\linewidth]{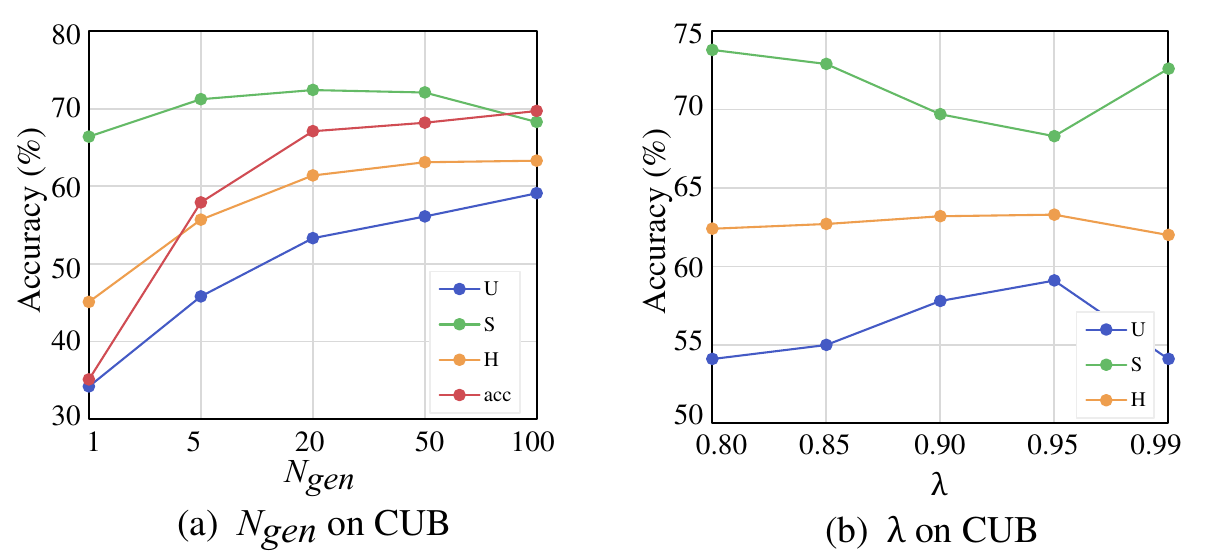}
        \caption{Hyper-parameter sensitivity. Take CUB dataset as an example, We explore (a) the effect of the number of generated images $N_{gen}$, and (b) the effect of calibration coefficient $\lambda$.}
        \label{fig:hyper}
    \end{center}
    \vspace{0mm}
\end{figure}

\noindent\textbf{Visualization of Generated Images.}
We present a visualization of the generated images at the identical step for our proposed DIG-ZSL and vanilla SD, as shown in Fig. \ref{fig:compare_imgs}. We make some promising observations: (1) The use of DCTs results in generated images that better align with text prompts. Specifically, for the Common Yellowthroat and Yellow-headed Blackbird, our approach successfully depicts their \textquotedblleft yellow throat\textquotedblright, and \textquotedblleft yellow head\textquotedblright, respectively. (2) DCTs improve the portrayal of details that the text does not provide. For example, the \textquotedblleft yellow bill\textquotedblright ~of the Red-legged Kittiwake. The last column displays a challenging case in which both the generated samples show significant distortion in details, such as the incomplete bird's feet. However, SD manages to render only part of the bill, while our DIG-ZSL successfully depicts the complete yellow bill. This result shows the effectiveness of our designs to generate realistic and informative images for unseen classes. More visualizations are in Appendix \ref{sec:appendB}.

\subsection{Hyper-Parameter Analysis}
\quad We investigate the impact of two different hyper-parameters on the CUB dataset: the number of generated images per unseen class, denoted as $N_{gen}$, and the calibration coefficient, $\lambda$. Fig. \ref{fig:hyper}(a) shows the results of varying the value of $N_{gen}$. As the number of samples increases, there is a corresponding upward trend in the metrics' values. However, the growth becomes stable when the number reaches a saturation point. Thus, enhancing the discrimination of the images is of paramount importance. Since generating numerous images per class time-consuming and resource-intensive, we simply set $N_{gen}$ to 100 for all the datasets. In Fig. \ref{fig:hyper}(b), we study the effect of $\lambda$ on CUB. In contrast to previous feature generation methods, which typically synthesize thousands of samples, our DIG-ZSL employs a calibration coefficient to calibrate the bias against seen classes. For datasets with a relatively larger number of images per class (\ie, AWA2, CUB, and FLO), we use a relative higher value (\eg, $\lambda$$=$0.95 for CUB). This result demonstrates that we select the reasonable hyper-parameters.
\section{Conclusion}
\label{sec:conclusion}
\quad In this paper, we propose a novel Discriminative Image Generation framework for ZSL (DIG-ZSL), which tackles ZSL in an image generation manner. By utilizing the semantic prototypes extracted from the text encoder of CLIP, we circumvent the dependence on human-annotated attributes. To generate diverse and informative images, we introduce a perspective of leveraging a pre-trained category discrimination model (CDM) to guide the optimization of discriminative class tokens (DCTs). The extensive experiments on four benchmark datasets demonstrate that our approach can facilitate ZSL task, achieving significant improvement compared with previous SOTA methods. We further visualize and analyze the qualitative results, which verify the capability of our DIG-ZSL to enhance text-to-image alignment and improve the portrayal of intricate detail. We hope to shed light on ZSL works in a new image generation paradigm.

{
\small
\bibliographystyle{ieeenat_fullname}
\bibliography{main}
}
\clearpage
\setcounter{page}{1}
\maketitlesupplementary

\appendix
Organization of the appendix:
\begin{itemize}
    \item Appendix \ref{sec:appendA}: More Implementation Details.
    \item Appendix \ref{sec:appendB}: More Visualizations.
\end{itemize}

\section{More Implementation Details}
\label{sec:appendA}
\noindent\textbf{Category Discrimination Model Training.}
The default setting is in Tab. \ref{tab:cdms}. We use the setting to learn an MLP.

\begin{table}[htbp]
  \centering
    \caption{Category discrimination model training setting.}
  \vspace{-1mm}
      \begin{tabular}{l|l}
        \hline
        Config & Value \\
        \hline
        optimizer & AdamW \\
        learning rate & 1e-3 \\
        optimizer momentum & $\beta_1$, $\beta_2$ = 0.5, 0.999 \\
        input size & $224\times{224}$ \\
        batch size & 64\\
        \hline
    \end{tabular}
      \label{tab:cdms}
       \vspace{1mm}
\end{table}

\noindent\textbf{Discriminative Class Token Learning.}
The default setting is in Tab. \ref{tab:dcts}. We use the setting to optimize the learning of discriminative class tokens (DCTs).

\begin{table}[htbp]
  \centering
    \caption{Discriminative class token learning setting.}
  \vspace{-1mm}
      \begin{tabular}{l|l}
        \hline
        Config & Value \\
        \hline
        optimizer & AdamW \\
        learning rate & 1.25e-3 \\
        weight decay & 0.01 \\
        optimizer momentum & $\beta_1$, $\beta_2$ = 0.9, 0.999 \\
        AdamW epsilon & 1e-8 \\
        width, height & 512, 512 \\
        downsamplingh factor & 8 \\
        guidance scale & 7 \\
        inference step & 50 \\
        early stop & 15 \\
        batch size & 5 \\
        number of prompts & 3 \\

        \hline
    \end{tabular}
      \label{tab:dcts}
       \vspace{1mm}
\end{table}

\noindent\textbf{Image Generation.}
Once we have obtained the Discriminative Class Token (DCT) for each unseen class, we integrate it into a text prompt to generate realistic and informative images. Specifically, we employ Stable Diffusion, using the input prompt \textquotedblleft A photo of $S_*$ [name]\textquotedblright, to produce these samples.

\noindent\textbf{Supervised Classifier Training.}
The default setting is the same as CDM, as shown in Tab. \ref{tab:dcts}. We utlize the setting to learn a classifier for ZSL.

\section{More Visualizations}
\label{sec:appendB}
\begin{figure}[htbp]
    \begin{center}
        \includegraphics[width=0.8\linewidth]{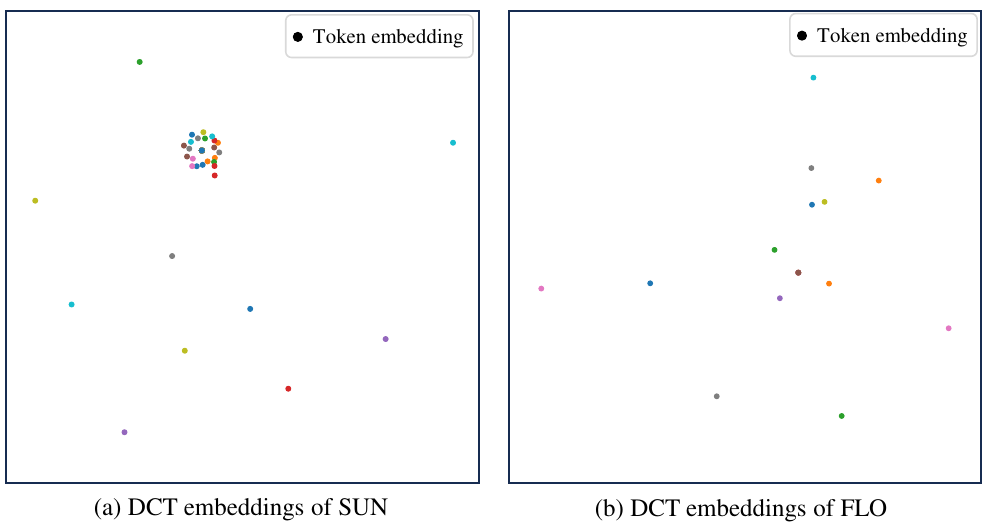}
        \caption{t-SNE visualization of the DCT embeddings for SUN and FLO. A dot denotes a DCT embedding for an unseen category. (Please zoom in for details.)}
        \label{fig:dct}
    \end{center}
\end{figure}

\noindent\textbf{Visualization of the DCT Embeddings.}
We further present the t-SNE visualization of the learned DCT embeddings for the SUN and FLO datasets, as depicted in Fig. \ref{fig:dct}. For the scene dataset SUN (see Fig. \ref{fig:dct}(a)), some tokens are observed to be close to each other. In contrast, for the FLO dataset (see Fig. \ref{fig:dct}(b)), the embeddings exhibit a scattered and discriminative pattern.

\begin{figure}[htbp]
    \begin{center}
        \includegraphics[width=0.9\linewidth]{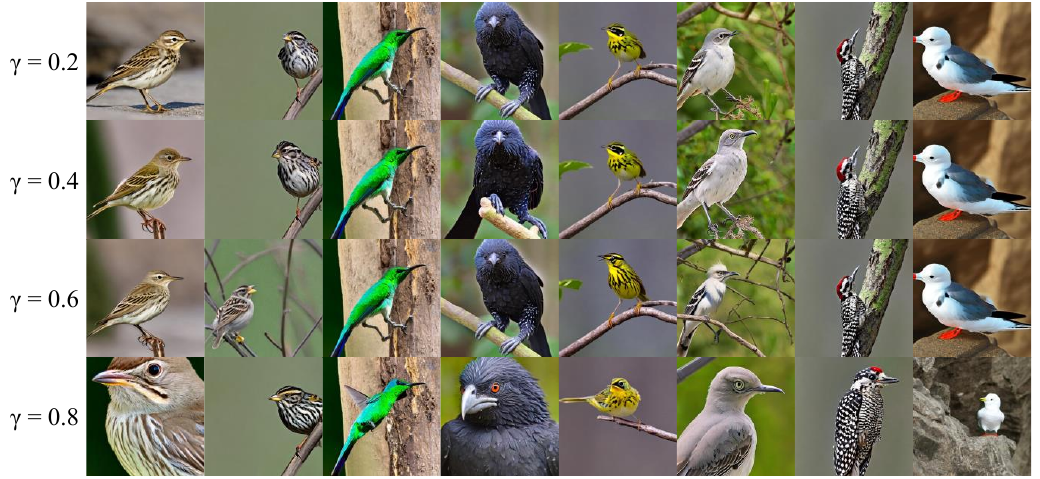}
        \caption{Visualization of different values of the threshold $\gamma$ on CUB dataset.}
        \label{fig:gamma}
    \end{center}
\end{figure}

\noindent\textbf{Visualization of Different Values of $\gamma$.}
We visualize the generated images at the same step with varying threshold values, as shown in Fig. \ref{fig:gamma}. For each threshold value, we train distinct DCTs. We can observe that an increase in $\gamma$ results in significant changes to the images. Thus, determining appropriate values for 
$\gamma$ is of great importance.

\begin{figure*}[t]
    \begin{center}
        \includegraphics[width=0.75\linewidth]{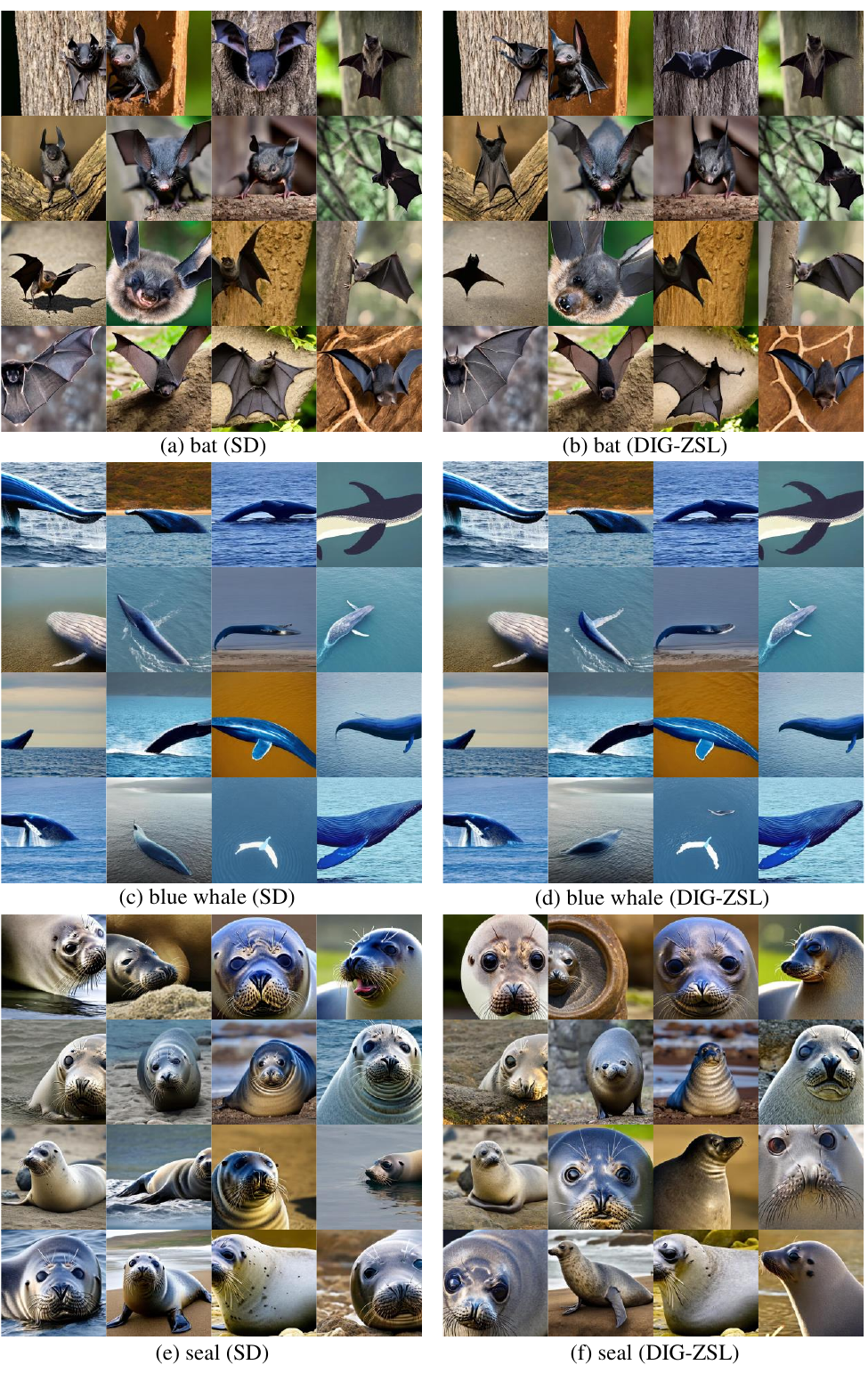}
        \caption{More visualizations on AWA2 dataset.}
        \label{fig:c0}
    \end{center}
\end{figure*}

\begin{figure*}[t]
    \begin{center}
        \includegraphics[width=0.75\linewidth]{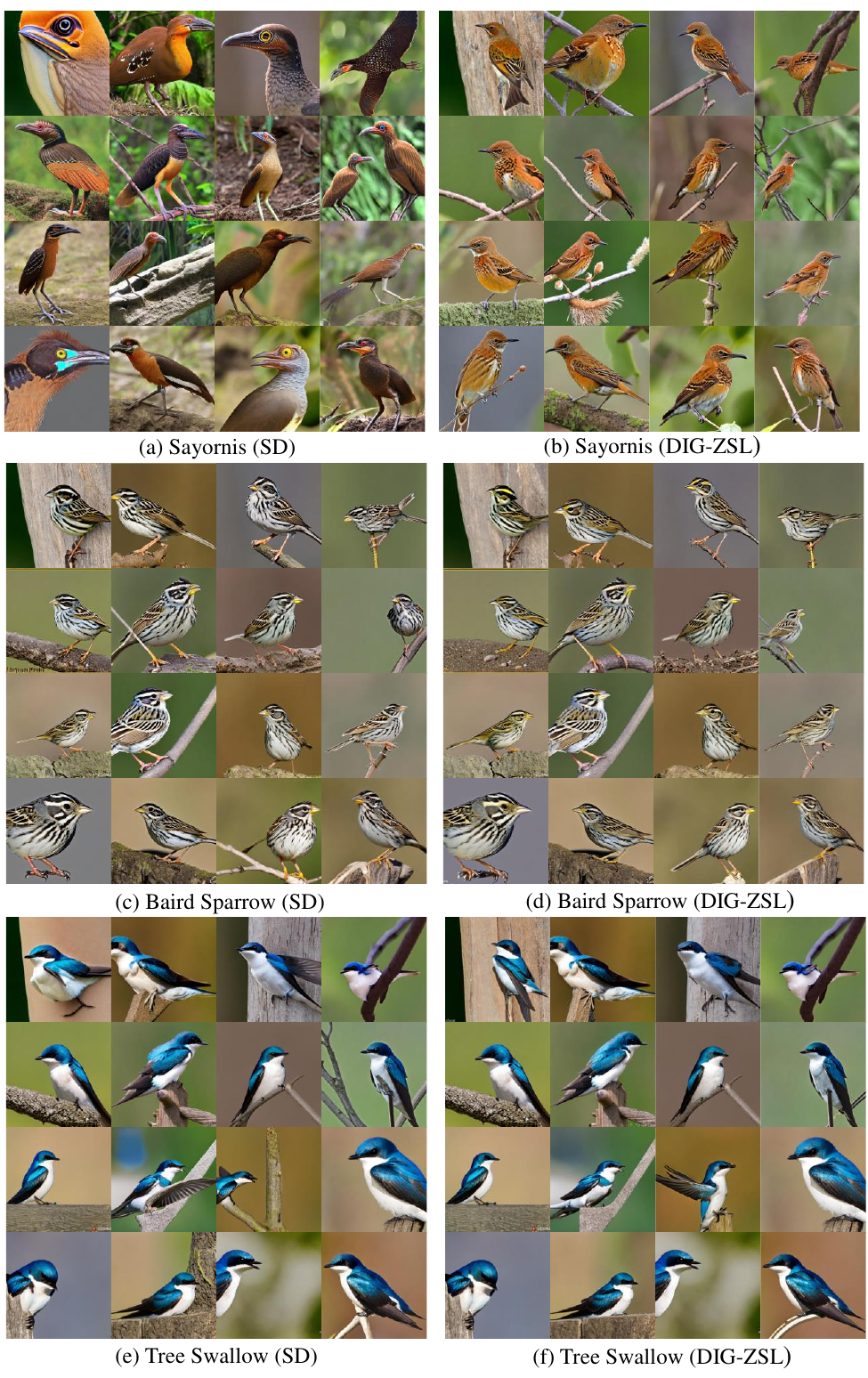}
        \caption{More visualizations on CUB dataset.}
        \label{fig:c1}
    \end{center}
\end{figure*}

\begin{figure*}[t]
    \begin{center}
        \includegraphics[width=0.75\linewidth]{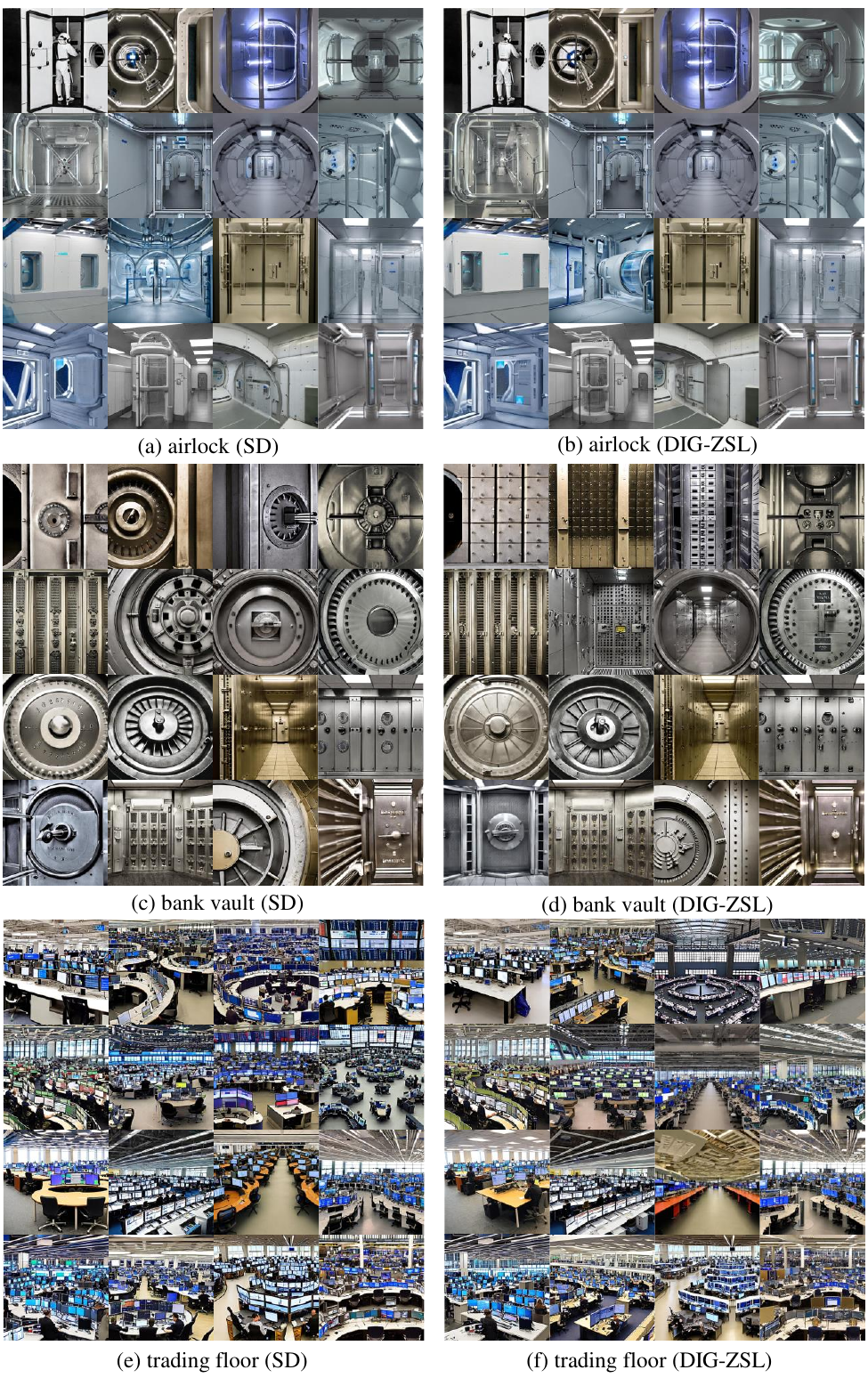}
        \caption{More visualizations on SUN dataset.}
        \label{fig:c2}
    \end{center}
\end{figure*}

\begin{figure*}[t]
    \begin{center}
        \includegraphics[width=0.75\linewidth]{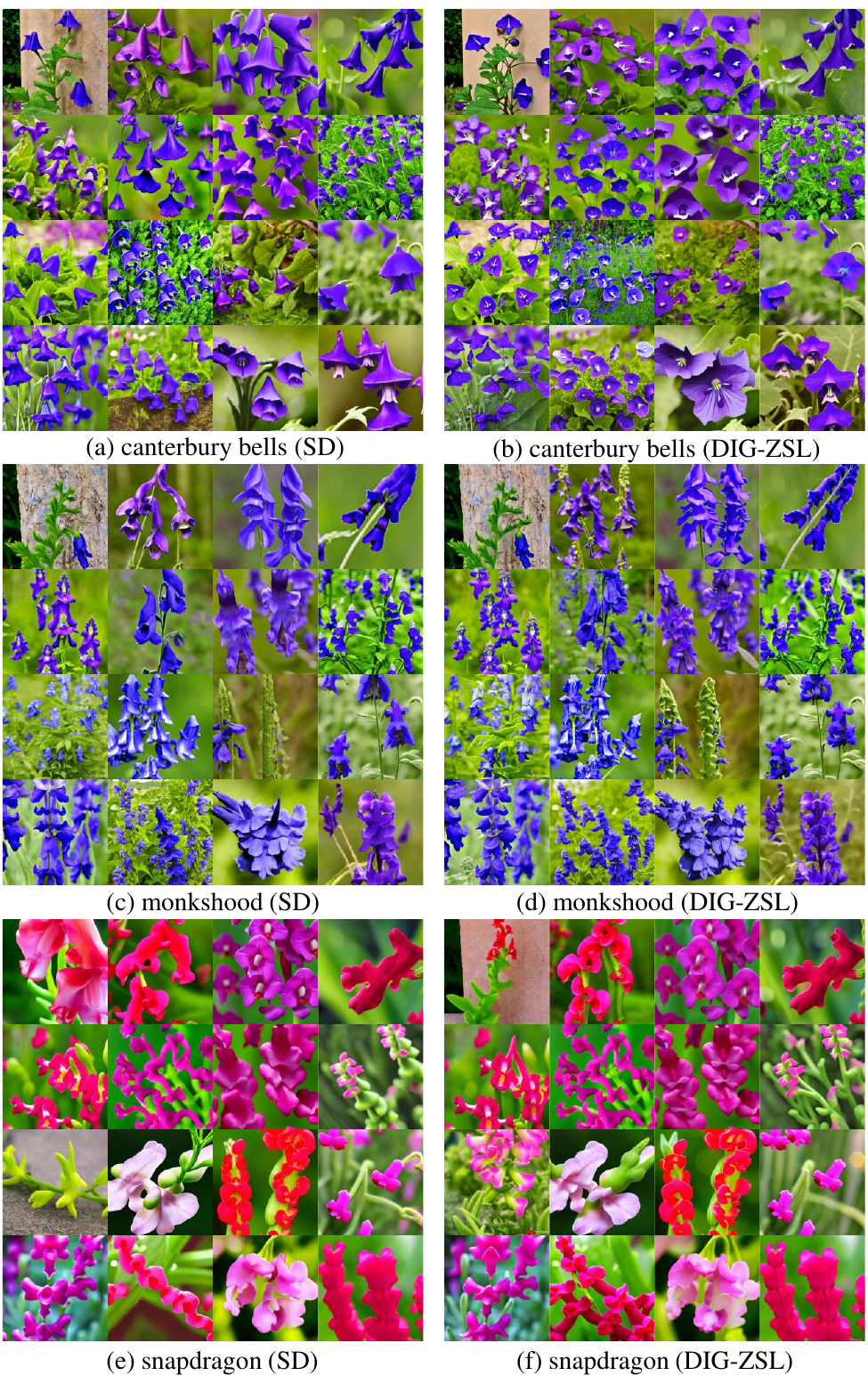}
        \caption{More visualizations on FLO dataset.}
        \label{fig:c3}
    \end{center}
\end{figure*}

\noindent\textbf{Visualization of the generated images.}
We further present more visualizations of the generated images of vanilla SD and our proposed DIG-ZSL. Fig. \ref{fig:c0} illustrates these comparisons on the AWA2 dataset, whereas Fig. \ref{fig:c1}, \ref{fig:c2} and \ref{fig:c3} depict the comparisons for the CUB, SUN, and FLO datasets, respectively. 

\end{document}